\documentclass{article}
\usepackage[preprint]{neurips_2024}

\usepackage[utf8]{inputenc} 
\usepackage[T1]{fontenc}    

\usepackage{hyperref} 
\usepackage{url}      
\usepackage{booktabs} 
\usepackage{amsfonts} 
\usepackage{nicefrac} 
\usepackage{microtype} 
\usepackage{xcolor}    
\usepackage{graphicx}
\usepackage{amsmath}
\usepackage{enumitem}  
\usepackage{float}   
\usepackage{subcaption}
\usepackage{authblk}
\usepackage{adjustbox}
\usepackage{multirow}
\usepackage{tikz}
\usetikzlibrary{arrows.meta, shapes.geometric, positioning, fit, backgrounds, decorations.pathmorphing, shadows.blur, patterns, calc}
\usepackage{fontawesome5}
\usepackage{graphicx}
\usepackage[utf8]{inputenc}
\usepackage[T1]{fontenc}
\usepackage{geometry}
\usepackage{xcolor}
\usepackage{listings}
\usepackage{amssymb} 
\definecolor{codegray}{gray}{0.95} 

\lstdefinestyle{PromptStyle}{
    backgroundcolor=\color{codegray},   
    basicstyle=\ttfamily\small,         
    breaklines=true,                    
    postbreak=\mbox{\textcolor{red}{$\hookrightarrow$}\space}, 
    frame=single,                       
    framerule=0pt,                      
    framexleftmargin=6pt,               
    framexrightmargin=6pt,
    framextopmargin=6pt,
    framexbottommargin=6pt,
    tabsize=2,                          
    showstringspaces=false,             
    commentstyle=\color{gray}\itshape,  
    keywordstyle=\bfseries,             
    morekeywords={PART 1, CONTEXT, PART 2, THE SPECIFIC PROBLEM, Premises to use, Conclusion to prove, PART 3, YOUR TASK, True, False, General Context, Task, Theorem, Pool of Premises, Question, Response Format, Proof Context & Rules, Shuffled Clauses, Required Output Format, Superposition Calculus, Resolution, Paramodulation},
    comment=[l]{--- EXPECTED OUTPUT ---} 
}

\lstdefinestyle{ModelOutputStyle}{
    backgroundcolor=\color{codegray},
    basicstyle=\ttfamily\scriptsize,
    breaklines=true,
    frame=tb, 
    framerule=0pt,
    framexleftmargin=6pt,
    tabsize=2,
    showstringspaces=false,
    numbers=left,
    numberstyle=\tiny\color{gray}
}
\newlength{\myMheight}
\newcommand{\github}{\includegraphics[height=\myMheight]{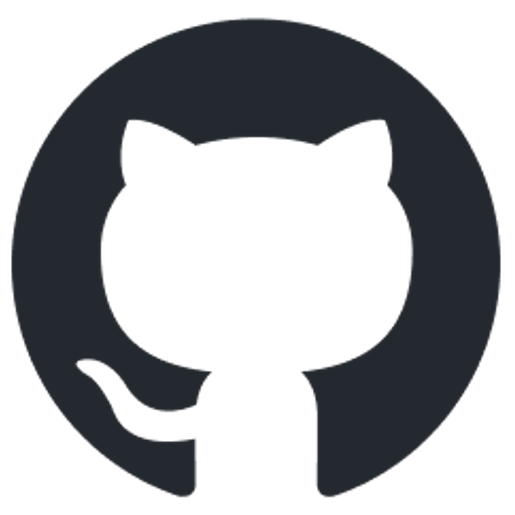}}
\newcommand{\hf}{\includegraphics[height=\myMheight]{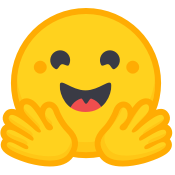}}

\title{Saturation-Driven Dataset Generation for LLM Mathematical Reasoning in the TPTP Ecosystem}

\author[]{Valentin Quesnel}
\author[]{Damien Sileo}
\affil[]{Univ. Lille, Inria, CNRS, Centrale Lille, UMR 9189 - CRIStAL, F-59000 Lille, France}
\affil[ ]{\texttt{valentinq2001@gmail.com, damien.sileo@inria.fr}}

\begin{document}

\maketitle

\begin{abstract}
The scarcity of high-quality, logically sound data is a critical bottleneck for advancing the mathematical reasoning of Large Language Models (LLMs). Our work confronts this challenge by turning decades of automated theorem proving research into a scalable data engine. Rather than relying on error-prone LLMs or complex proof-assistant syntax like Lean/Isabelle, our framework leverages E-prover's saturation capabilities on the vast TPTP axiom library to derive a massive, guaranteed-valid corpus of theorems. Our pipeline is principled and simple: saturate axioms, filter for "interesting" theorems, and generate tasks. With no LLMs in the loop, we eliminate factual errors by construction. This purely symbolic data is then transformed into three difficulty-controlled challenges: entailment verification, premise selection, and proof reconstruction. Our zero-shot experiments on frontier models reveal a clear weakness: performance collapses on tasks requiring deep, structural reasoning. Our framework provides both the diagnostic tool to measure this gap and a scalable source of symbolic training data to address it. We make the code and data publicly available\footnote{Code and data available at:\\ \github : \url{https://github.com/sileod/reasoning_core} \\
\hf : \url{https://hf.co/datasets/reasoning-core/rc1} }

\end{abstract}

\section{Introduction}

Advancing the mathematical reasoning capabilities of Large Language Models (LLMs) is a central goal in AI, yet their performance in rigorous, multi-step logical deduction remains a critical weakness \citep{li2024}. This deficiency is largely attributed to a foundational issue: the extreme scarcity of high-quality training data \citep{wang2020, xin2024a}. Unlike domains rich with web-scale text or images, the corpus of formalized mathematical proofs is small, expensive to create, and requires specialized human expertise, making it impractical to crowdsource at scale \citep{wang2020, Trinh2024}. This data bottleneck severely hampers the training of more capable mathematical reasoners, motivating approaches that are logically grounded, resource-aware, and reproducible. 

Recently, a promising line of research has focused on synthetic data generation to overcome this scarcity \citep{yin2025, wang2020}. However, many of these approaches either generate problems by prompting other LLMs, which carries a significant risk of introducing factual inconsistencies that require complex external verification in Interactive Theorem Provers (ITPs) like Lean or Isabelle \citep{xin2024a, wu2022}, or generate random logical formulas that may lack mathematical relevance \citep{zombori2019}.

In this paper, we propose a different paradigm. Our framework is built in symbiosis with the mature automated theorem proving ecosystem. We use the saturation-based theorem prover E \citep{schulz2002} not as a solver, but as a \textbf{generative engine} to exhaustively derive logical consequences from the rich TPTP axiom library \citep{sutcliffe2017}. This purely symbolic approach ensures the mathematical validity of every generated theorem by construction. After curating these results for "interestingness" with AGInTRater \citep{puzis2006}, we deconstruct the process of theorem proving into three complementary facets of logical reasoning, each embodied by a distinct task with controllable complexity:
\begin{enumerate}
    \item \textbf{Conjecture Entailment:} A true/false task testing deductive verification.
    \item \textbf{Minimal Premise Selection:} A distractor-based task requiring the identification of the minimal sufficient set of hypotheses
    \item \textbf{Proof Graph Reconstruction:} A structural reasoning task that involves ordering shuffled proof steps.
\end{enumerate}

Our contribution is not a static dataset but a generative mechanism that produces varied and complex tasks on demand. It provides a \textbf{granular method for evaluating LLM reasoning depth}, isolating the pure logical capabilities of models from the ambiguities of natural language.

\section{Related Work}

Our work addresses the critical need for high-quality training data in automated theorem proving. We position our contribution with respect to two dominant paradigms: LLM-based autoformalization and procedural generation within Interactive Theorem Provers (ITPs).

\paragraph{Synthetic Data via LLM-driven Pipelines.} A major line of research leverages LLMs to create new training instances, primarily through autoformalization—translating natural language problems into formal systems like Lean \citep{xin2024a, wu2022}. As seen in projects like TheoremLlama \citep{wang2024}, this paradigm can generate large datasets. However, this "LLM-in-the-loop" approach faces significant drawbacks. It relies on the very models it seeks to improve and, because LLM outputs are not guaranteed to be sound, necessitates a computationally expensive "generate-then-verify" cycle using an Interactive Theorem Prover (ITP). This dependency not only confines the generation process to a specific and complex proof-assistant syntax (e.g., Lean/Isabelle) but also suffers from the high cost of rejection sampling due to frequent invalid proposals. Furthermore, being trained on human data, LLMs tend to generate problems with a generative bias, favoring familiar structures over more "exotic" or diverse logical paths. Our framework avoids these issues entirely: by using saturation within the standardized TPTP ecosystem, we guarantee validity by construction, operate independently of any proof-assistant syntax, and systematically explore the deductive space for maximal diversity.

\paragraph{Procedural Generation and Proof Automation in ITPs.} A second stream of work focuses on procedural generation and proof automation within ITPs. Similar to our work, \citet{Ayg_n2020} also generate synthetic theorems from TPTP axioms using a forward proposer. However, their method uses a simple linear resolution strategy with random choices, which does not guarantee the exploration of complex or diverse logical structures. Our approach is more systematic: by using a full saturation engine (E-prover), we exhaustively explore a part of the deductive closure of the axioms, a process limited in practice by a prover timeout. Furthermore, we curate this vast output with the established interest metrics of AGInTRater \citep{puzis2006} to focus on non-trivial theorems, a step absent in simpler random-walk generation.

This contrast is also visible when comparing to procedural generation in ITPs like Metamath or Lean \citep{wang2020, yin2025}. These methods often start from existing human-written proofs to guide their search. Our approach is more foundational, requiring only axioms. Moreover, while ITPs are powerful, they are fundamentally designed for human-guided, goal-directed proof, not for exhaustive, open-ended saturation. Tools like Sledgehammer and Duper \citep{clune2024} are powerful goal-directed solvers, but not inherently generative. By using a tool purpose-built for saturation (E-prover), our framework is architecturally better suited for the large-scale, systematic generation we propose.

\paragraph{Task Formulation for Evaluation.} While most data generation efforts focus on creating (theorem, proof) pairs for training provers \citep{lample2022, Ayg_n2020}, a deeper evaluation of reasoning requires more granular tasks. Premise selection, for instance, has been identified as a key bottleneck \citep{irving2016}, yet is often treated as a preliminary step rather than a standalone evaluation. Our contribution is therefore not just a data generation method, but the formulation of a structured suite of complementary tasks—entailment verification, minimal premise selection, and proof graph reconstruction. By deconstructing the theorem-proving process, a concept explored in systems like TRAIL \citep{crouse2021} which analyzes proof states, we provide a robust framework for diagnosing the specific logical capabilities and failure modes of LLMs.

\section{The Framework}

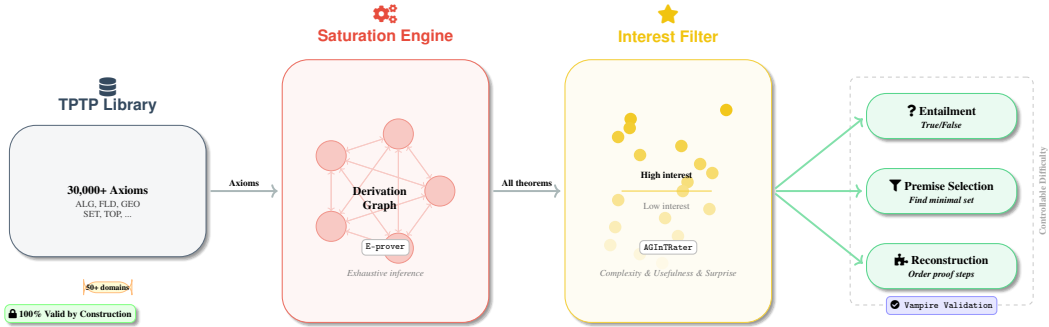
\begin{figure*}[h!]
\centering
\resizebox{\textwidth}{!}{
\begin{tikzpicture}[
    /utils/exec={
        \definecolor{tptpcolor}{RGB}{52, 73, 94}
        \definecolor{saturatecolor}{RGB}{231, 76, 60}
        \definecolor{ratecolor}{RGB}{241, 196, 15}
        \definecolor{taskcolor}{RGB}{46, 204, 113}
        \definecolor{arrowcolor}{RGB}{149, 165, 166}
    },
    mainBox/.style={
        rectangle,
        rounded corners=20pt,
        thick,
        align=center,
        text width=5cm,
        blur shadow={shadow blur steps=5}
    },
    inputBox/.style={
        mainBox,
        minimum height=3.5cm,
        draw=tptpcolor!80, fill=tptpcolor!5
    },
    processBox/.style={
        mainBox, minimum width=5.5cm, minimum height=7cm,
    },
    taskBox/.style={
        rectangle, rounded corners=15pt, thick,
        draw=taskcolor!80, fill=taskcolor!10,
        minimum width=4cm, minimum height=1.2cm,
        text centered, align=center, font=\footnotesize\bfseries,
        blur shadow={shadow blur steps=3}
    },
    toolLabel/.style={
        rectangle, rounded corners=3pt,
        fill=white, draw=gray!50, thick,
        font=\scriptsize\ttfamily, inner sep=3pt
    },
    metricLabel/.style={
        rectangle, rounded corners=10pt,
        fill=yellow!20, draw=orange!50,
        font=\tiny\bfseries, inner sep=2pt
    },
    arrow/.style={
        ->, thick, draw=arrowcolor!80, 
        shorten >=3pt, shorten <=3pt,
        line width=1.5pt
    },
    dashedBox/.style={
        rectangle, dashed, draw=gray!40, thick,
        rounded corners=5pt
    },
    iconStyle/.style={font=\Large, text=white},
    titleStyle/.style={font=\large\bfseries\sffamily}
]

\node[inputBox] (tptp) {};
\node[iconStyle, text=tptpcolor] at ([yshift=1cm]tptp.north) {\faDatabase};
\node[titleStyle, text=tptpcolor] at ([yshift=0.5cm]tptp.north) {TPTP Library};
\node[font=\small\bfseries] at (tptp.center) {30,000+ Axioms};
\node[font=\scriptsize, text=gray!150] at ([yshift=-0.5cm]tptp.center) {
    \begin{tabular}{c}
    ALG, FLD, GEO\\
    SET, TOP, ...
    \end{tabular}
};
\node[metricLabel] at ([yshift=-0.8cm]tptp.south) {50+ domains};

\node[processBox, draw=saturatecolor!80, fill=saturatecolor!5, right=2cm of tptp] (saturate) {};
\node[iconStyle, text=saturatecolor] at ([yshift=1.2cm]saturate.north) {\faCogs};
\node[titleStyle, text=saturatecolor] at ([yshift=0.6cm]saturate.north) {Saturation Engine};

\node[inner sep=0] at (saturate.center) {
    \begin{tikzpicture}[scale=0.8]
        \foreach \i in {0,72,144,216,288} {
            \node[circle, fill=saturatecolor!30, draw=saturatecolor!60, minimum size=8mm] (n\i) at (\i:2) {};
        }
        \foreach \i in {0,72,144,216,288} {
            \foreach \j in {0,72,144,216,288} {
                \draw[saturatecolor!30, ->] (n\i) -- (n\j);
            }
        }
        \node at (0,0) [font=\footnotesize\bfseries] {Derivation};
        \node at (0,-0.5) [font=\footnotesize\bfseries] {Graph};
    \end{tikzpicture}
};

\node[toolLabel] at ([yshift=-1.5cm]saturate.center) {E-prover};
\node[font=\scriptsize\itshape, text=gray] at ([yshift=-2.2cm]saturate.center) {Exhaustive inference};

\node[processBox, draw=ratecolor!80, fill=ratecolor!5, right=2cm of saturate] (rate) {};
\node[iconStyle, text=ratecolor] at ([yshift=1.2cm]rate.north) {\faStar};
\node[titleStyle, text=ratecolor] at ([yshift=0.6cm]rate.north) {Interest Filter};

\node at (rate.center) {
    \begin{tikzpicture}[scale=0.8]
        \foreach \i in {1,...,19} {
            \pgfmathsetmacro{\opacity}{100-\i*5}
            \node[circle, fill=ratecolor!\opacity, minimum size=3mm] at ({rand*2}, {5-(\i*0.3)}) {};
        }
        \draw[thick, ratecolor!80] (-1.5, 2) -- (1.5, 2);
        \node[font=\scriptsize\bfseries] at (0, 2.5) {High interest};
        \node[font=\scriptsize, text=gray] at (0, 1.5) {Low interest};
    \end{tikzpicture}
};

\node[toolLabel] at ([yshift=-1.5cm]rate.center) {AGInTRater};
\node[font=\scriptsize\itshape, text=gray] at ([yshift=-2.2cm]rate.center) {Complexity \& Usefulness \& Surprise};

\node[taskBox, right=2.5cm of rate, yshift=2cm] (task1) {
\faQuestion\ Entailment\\
\scriptsize\textit{True/False}
};
\node[taskBox, right=2.5cm of rate] (task2) {
\faFilter\ Premise Selection\\
\scriptsize\textit{Find minimal set}
};
\node[taskBox, right=2.5cm of rate, yshift=-2cm] (task3) {
\faPuzzlePiece\ Reconstruction\\
\scriptsize\textit{Order proof steps}
};
    
\node[rotate=90, font=\scriptsize\bfseries, text=gray!70] at ([xshift=0.7cm]task2.east) {Controllable Difficulty};

\node[dashedBox, fit=(task1)(task2)(task3), inner sep=12pt] (validation_box) {};
\node[toolLabel, fill=blue!10, draw=blue!40] at (validation_box.south) {\faCheckCircle\ Vampire Validation};

\draw[arrow] (tptp.east) -- (saturate.west) 
    node[midway, above, font=\scriptsize\bfseries] {Axioms};
\draw[arrow] (saturate.east) -- (rate.west)
    node[midway, above, font=\scriptsize\bfseries] {All theorems};

\draw[arrow, taskcolor!60] (rate.east) -- (task1.west);
\draw[arrow, taskcolor!60] (rate.east) -- (task2.west);
\draw[arrow, taskcolor!60] (rate.east) -- (task3.west);

\node[rectangle, rounded corners, fill=green!10, draw=green!50, 
      font=\scriptsize\bfseries, inner sep=3pt, blur shadow={shadow blur steps=3}] 
      at ([yshift=-1.5cm, xshift=-1cm]tptp.south) 
      {\faLock\ 100\% Valid by Construction};

\end{tikzpicture}}
\caption{From TPTP axioms to reasoning tasks through symbolic saturation. 
    The pipeline guarantees mathematical validity by construction, with E-prover generating exhaustive derivation graphs, 
    AGInTRater filtering for mathematical interest, and Vampire validating all outputs. 
    Each stage is fully deterministic and reproducible, producing unlimited tasks with controllable difficulty.}
\label{fig:framework_overview}
\end{figure*}

Our framework is a multi-stage, fully symbolic pipeline that transforms foundational axioms into a rich repository of structured reasoning tasks. The process is designed to be automated and logically sound by construction, leveraging a synergy of established automated reasoning tools.

\subsection{Stage 1: Exhaustive Generation via Saturation}

The core of our method is the generation of a comprehensive directed acyclic graph (DAG) of logical derivations. We begin by selecting an axiom file from a specific domain within the TPTP library, expressed in clausal normal form (CNF).

For this generative step, we specifically chose the \textbf{E theorem prover}. While many ATPs are optimized for goal-directed search, E-prover offers exceptional flexibility and can be configured to run in a pure \textbf{saturation mode} without a specific conjecture to prove. Saturation is a proof procedure that exhaustively applies inference rules of the superposition calculus (like resolution and paramodulation) to explore the deductive closure of the axiom set, a process that in practice is bounded by computational limits, such as a timeout, due to the combinatorial explosion of clauses. We configure E-prover to output a full proof graph, which meticulously records every derived clause and its parent(s). This process yields a vast graph containing a systematic enumeration of the logical consequences of the initial axiom set.

\subsection{Stage 2: Curation with Interest Metrics}

The saturation process is exhaustive by design and often produces a combinatorial explosion of clauses, many of which are trivial, redundant, or overly complex. To focus on mathematically meaningful theorems, we introduce a curation stage where the entire derivation graph is analyzed to identify the most "interesting" results.

For this, we leverage AGInTRater \citep{puzis2006}, a system that evaluates logical formulas not in isolation, but within the context of their derivation. AGInTRater assigns a score to each non-axiom clause based on a combination of heuristic metrics designed to emulate human judgment. These metrics include:

\paragraph{Complexity and Weight} Measures the structural size of a formula, penalizing overly cumbersome statements.
\paragraph{Surprisingness:} Evaluates new relationships between concepts by measuring how infrequently function and predicate symbols co-occur in the initial axioms.
\paragraph{Usefulness:} Assesses a theorem's utility as a lemma by calculating the ratio of its "interesting" descendants to its total descendants within the derivation graph.

By providing the entire graph to AGInTRater, we obtain a contextualized score for each potential theorem. This crucial step prunes the vast space of logical consequences, ensuring that the tasks generated downstream are based on non-trivial and structurally relevant mathematical statements.

We emphasize that AGInTRater provides a heuristic score. We use it to prioritize non-trivial theorems in a principled way, without claiming alignment with human judgments of mathematical interest. In our experiments we simply retain the top-scoring items per domain and depth setting.

\subsection{Stage 3: Task Formulation and Validation}

The filtered graph is modeled using the NetworkX library. For each task generated, we use the \textbf{Vampire theorem prover} \citep{kovacs2013} as our ground-truth oracle. We chose Vampire for this validation role due to its state-of-the-art performance and speed, consistently ranking as a top prover in the CASC competitions. Its efficiency allows us to rapidly validate the thousands of logical entailment queries needed to generate our benchmark tasks with 100\% confidence up to Vampire timeouts .

\section{Generated Reasoning Tasks}

From the curated and structured derivation graph, we deconstruct the process of theorem proving into three complementary facets of logical reasoning, each embodied by a distinct task of increasing complexity.

\subsection{Task 1: Conjecture Entailment Verification}

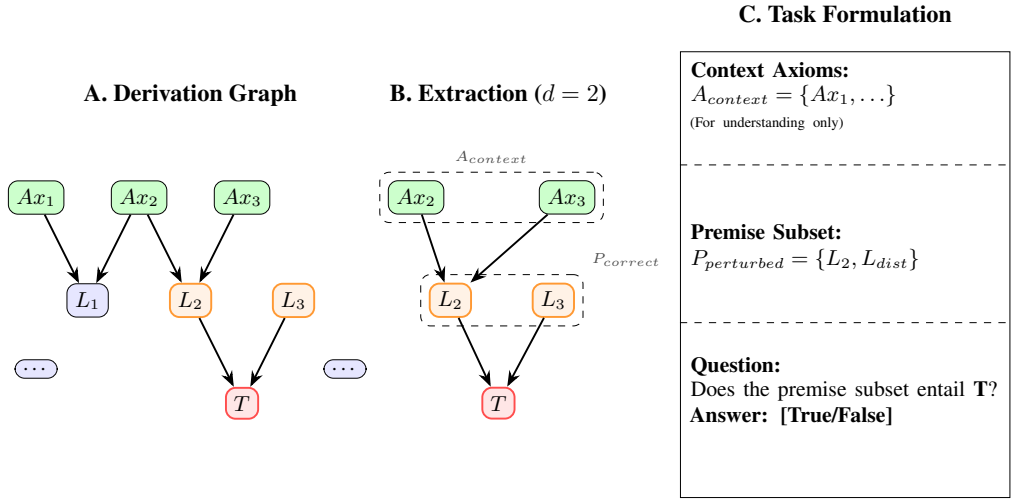
\begin{figure}[h!]
    \centering 
    \resizebox{\textwidth}{!}{
    \tikzstyle{clause}  = [rectangle, rounded corners, draw=black, fill=blue!10, font=\footnotesize, align=center]
\tikzstyle{distractor}  = [rectangle, rounded corners, draw=black, fill=orange!10, font=\footnotesize, align=center]
\tikzstyle{axiom}   = [rectangle, rounded corners, draw=black, fill=green!20, font=\footnotesize, align=center]
\tikzstyle{target}  = [rectangle, rounded corners, draw=red!70, thick, fill=red!10, font=\footnotesize, align=center]
\tikzstyle{premise} = [rectangle, rounded corners, draw=orange!80, thick, fill=orange!10, font=\footnotesize, align=center]
\tikzstyle{panel_label} = [font=\bfseries]
\tikzstyle{box_label}   = [font=\tiny\itshape, text=black!70]
\tikzstyle{arrow}       = [->, >=Stealth, thick]
\tikzstyle{process_arrow} = [->, decorate, decoration={snake, amplitude=.3mm, segment length=1.5mm, post length=1mm}, thick, draw=orange!80]

\begin{tabular}{@{}ccc@{}}
\begin{minipage}{0.32\textwidth}
    \centering
    \begin{tikzpicture}[node distance=1.5cm and 0.8cm, >=Stealth]
        \node[panel_label] (label_a) at (2.25, 4) {A. Derivation Graph};
        \node[axiom] (ax1) at (0, 2.5) {$Ax_1$};
        \node[axiom] (ax2) at (1.5, 2.5) {$Ax_2$};
        \node[axiom] (ax3) at (3, 2.5) {$Ax_3$};
        \node[clause] (c1) at (0.75, 1) {$L_1$};
        \node[premise] (c2) at (2.25, 1) {$L_2$};
        \node[premise] (c3) at (3.75, 1) {$L_3$};
        \node[target] (t) at (3, -0.5) {$T$};
        \node[clause] (c4) at (4.5, 0) {\dots};
        \node[clause] (c5) at (0, 0) {\dots};
        \draw[arrow] (ax1) -- (c1); \draw[arrow] (ax2) -- (c1);
        \draw[arrow] (ax2) -- (c2); \draw[arrow] (ax3) -- (c2);
        \draw[arrow] (c2) -- (t); \draw[arrow] (c3) -- (t);
    \end{tikzpicture}
\end{minipage}
&
\begin{minipage}{0.32\textwidth}
    \centering
    \begin{tikzpicture}[node distance=1.5cm and 0.8cm, >=Stealth]
        \node[panel_label] (label_b) at (1.2, 4) {B. Extraction ($d=2$)};
        
        \node[axiom] (b_ax2) at (0, 2.5) {$Ax_2$};
        \node[axiom] (b_ax3) at (2.2, 2.5) {$Ax_3$};
        
        \node[premise] (b_c2) at (0.5, 1) {$L_2$};
        \node[premise] (b_c3) at (2.0, 1) {$L_3$};
        
        \node[target]  (b_t)  at (1.2, -0.5) {$T$};
        
        \draw[arrow] (b_ax2) -- (b_c2); \draw[arrow] (b_ax3) -- (b_c2);
        \draw[arrow] (b_c2) -- (b_t);  \draw[arrow] (b_c3) -- (b_t);
        
        \node[fit=(b_c2)(b_c3), draw, rounded corners, dashed, label={[box_label, overlay]north east:$P_{correct} $}] {};
        
        \node[fit=(b_ax2)(b_ax3), draw, rounded corners, dashed, label={[box_label]above:$A_{context}$}] {};
    \end{tikzpicture}
\end{minipage}
&
\begin{minipage}{0.32\textwidth}
    \centering
    \begin{tikzpicture}
        \node[panel_label] (label_c) at (0, 4) {C. Task Formulation};
        \node[rectangle, draw, minimum width=4.8cm, minimum height=6.5cm] (llm_box) at (0, 0.2) {};
        \node[align=left, text width=4.5cm] at (0, 2.8) {\small \textbf{Context Axioms:} \\ \footnotesize $A_{context} = \{Ax_1, \ldots\}$ \\ \tiny (For understanding only)};
        \draw[dashed, thin] (llm_box.west |- 0, 1.8) -- (llm_box.east |- 0, 1.8);
        \node[align=left, text width=4.5cm] at (0, 0.6) {\small \textbf{Premise Subset:} \\ \footnotesize $P_{perturbed} = \{L_2, L_{dist}\}$};
        \draw[dashed, thin] (llm_box.west |- 0, -0.5) -- (llm_box.east |- 0, -0.5);
        \node[align=left, text width=4.5cm] at (0, -1.5) {\small \textbf{Question:} \\ \footnotesize Does the premise subset entail \textbf{T}? \\ \footnotesize \textbf{Answer: [True/False]}};
    \end{tikzpicture}
\end{minipage}
\end{tabular}
    }
    
    \caption{Generation pipeline for a Conjecture Entailment Verification task. A: We start from a curated derivation graph. B: We extract a minimal set of lemmas $P_{correct}$ at depth $d$. C: We form a perturbed set $P_{perturbed}$ and ask if it entails $T$.}

    \label{Entailement Task}
\end{figure}

\paragraph{Context and Premise Extraction:} We select an interesting theorem $T$ from the curated graph. We perform two backward traversals from $T$:
\begin{itemize}
    \item \textbf{Context Axioms:} We trace the derivation all the way back to the foundational axioms to identify the complete set of axioms, $A_{\text{context}}$, required for the proof. These axioms provide the necessary theoretical background (e.g., definitions of operators and predicates) for a model to understand the problem's domain.
    \item \textbf{Immediate Premises:} We perform a second traversal up to a predefined proof depth $d$. This yields a set of intermediate lemmas, $P_{\text{correct}}$, that are sufficient to prove $T$ in $d$ steps.
\end{itemize}

\paragraph{Controlled Premise Set Perturbation:} We then create a perturbed premise set $P_{\text{perturbed}}$ by applying a controlled number of transformations, $k$, to $P_{\text{correct}}$. These transformations include adding, removing, or replacing clauses from the original derivation graph. The number of perturbations $k$ is another parameter to modulate the task's difficulty.

\paragraph{Task Formulation and Ground Truth:} The LLM is presented with the context axioms $A_{\text{context}}$, the perturbed premise set $P_{\text{perturbed}}$, and the theorem $T$. The prompt explicitly instructs the model that the context axioms are for background understanding only, and that the entailment question—"Does this specific set of premises entail the theorem?"—must be answered using only the statements in $P_{\text{perturbed}}$. The ground truth (True or False) for this specific entailment is predetermined by Vampire.

By providing this contextual background while strictly delimiting the premises allowed for the proof, we test the model's ability to distinguish between general domain knowledge and the specific hypotheses relevant to a particular deductive task. Modulating both \texttt{proof depth} and perturbations allows us to precisely map the limits of this core reasoning skill.

\subsection{Task 2: Minimal Premise Selection}

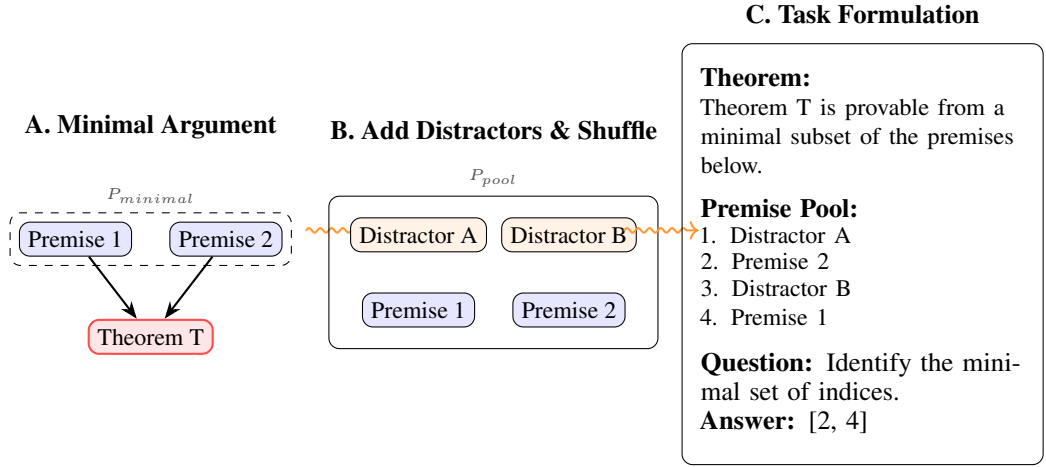
\begin{figure}[h!]
    \centering 
    
    \tikzstyle{clause}  = [rectangle, rounded corners, draw=black, fill=blue!10, font=\footnotesize, align=center]
\tikzstyle{distractor}  = [rectangle, rounded corners, draw=black, fill=orange!10, font=\footnotesize, align=center]
\tikzstyle{axiom}   = [rectangle, rounded corners, draw=black, fill=green!20, font=\footnotesize, align=center]
\tikzstyle{target}  = [rectangle, rounded corners, draw=red!70, thick, fill=red!10, font=\footnotesize, align=center]
\tikzstyle{premise} = [rectangle, rounded corners, draw=orange!80, thick, fill=orange!10, font=\footnotesize, align=center]
\tikzstyle{panel_label} = [font=\bfseries]
\tikzstyle{box_label}   = [font=\tiny\itshape, text=black!70]
\tikzstyle{arrow}       = [->, >=Stealth, thick]
\tikzstyle{process_arrow} = [->, decorate, decoration={snake, amplitude=.3mm, segment length=1.5mm, post length=1mm}, thick, draw=orange!80]

\begin{minipage}{0.28\textwidth}
    \centering
    \begin{tikzpicture}
        \node[panel_label] (label_a) at (0, 2.8) {A. Minimal Argument};
        \node[clause] (p1) at (-1, 1.3) {Premise 1};
        \node[clause] (p2) at ( 1, 1.3) {Premise 2};
        \node[target] (t_a) at (0, 0) {Theorem T};
        \draw[arrow] (p1) -- (t_a); \draw[arrow] (p2) -- (t_a);
        \node[fit=(p1)(p2), draw, rounded corners, dashed, label={[box_label]above:$P_{minimal}$}] (panelAfit) {};
    \end{tikzpicture}
\end{minipage}
\begin{minipage}{0.01\textwidth}
    \centering
    \begin{tikzpicture} \draw[process_arrow] (0,0) -- (0.8,0); \end{tikzpicture}
\end{minipage}
\begin{minipage}{0.28\textwidth}
    \centering
    \begin{tikzpicture}
        \node[panel_label] (label_b) at (0, 3.2) {B. Add Distractors \& Shuffle};
        \node[clause] (b_p1) at (-1, 0.8) {Premise 1};
        \node[clause] (b_p2) at ( 1, 0.8) {Premise 2};
        \node[distractor] (d1) at (-1, 1.8) {Distractor A};
        \node[distractor] (d2) at ( 1, 1.8) {Distractor B};
        \node[fit=(b_p1)(b_p2)(d1)(d2), draw, rounded corners, label={[box_label]above:$P_{pool}$}, inner sep=8pt] (pool_box) {};
    \end{tikzpicture}
\end{minipage}
\begin{minipage}{0.05\textwidth}
    \centering
    \begin{tikzpicture} \draw[process_arrow] (0,0) -- (1,0); \end{tikzpicture}
\end{minipage}
\begin{minipage}{0.28\textwidth}
    \centering
    \begin{tikzpicture}
        \node[panel_label] (label_c) at (0, 3.2) {C. Task Formulation};
        \node[rectangle, draw, rounded corners, minimum width=4.8cm, minimum height=5.6cm] (llm_box) at (0,0) {};
        
        \node[align=left, text width=4.3cm, anchor=north] at ([yshift=-2mm]llm_box.north) {
            \textbf{Theorem:}\\
            \footnotesize Theorem T is provable from a minimal subset of the premises below.
        };
        \node[align=left, text width=4.3cm, anchor=center] at ([yshift=-1mm]llm_box.center) {
            \textbf{Premise Pool:}\\
            \footnotesize 1. Distractor A\\ \footnotesize 2. Premise 2\\
            \footnotesize 3. Distractor B\\ \footnotesize 4. Premise 1
        };
        \node[align=left, text width=4.3cm, anchor=south] at ([yshift=3mm]llm_box.south) {
            \textbf{Question:} Identify the minimal set of indices.\\
            \textbf{Answer:} [2, 4]
        };
    \end{tikzpicture}
\end{minipage}
    
    \caption{Generation pipeline for a Minimal Premise Selection task. A: theorem with minimal premises $P_{minimal}$. B: embed into a larger pool $P_{pool}$ with distractors and shuffle. C: the LLM receives the shuffled, numbered pool and must return the indices of $P_{minimal}$.}

    \label{Premises Selection Task}
\end{figure}

This task targets a more advanced skill: identifying the necessary and sufficient hypotheses for a proof from a noisy context. The difficulty is controlled by both the logical depth of the argument and the amount of distracting information.

\paragraph{Generation Process:}
\begin{enumerate}
    \item \textbf{Context and Theorem Selection:} We select an interesting theorem $T$ from the curated graph and extract its foundational \textbf{context axioms}, $A_{\text{context}}$.
    \item \textbf{Minimal Premise Identification at Depth $d$:} We identify a minimal set of premises, $P_{\text{minimal}}$, required to prove $T$. To control the argument's complexity, we first extract a sufficient premise set, $P_{\text{sufficient}}$, by performing a backward traversal from $T$ up to a predefined \textbf{proof depth} $d$. We then apply an iterative pruning algorithm on $P_{\text{sufficient}}$, using \textbf{Vampire} to guarantee minimality.
    \item \textbf{Premise Pool Creation with $k$ Distractors:} We create a larger pool of premises, $P_{\text{pool}}$, by embedding $P_{\text{minimal}}$ within a set of $k$ \textbf{distractor clauses}. The number of distractors, $k$, is a parameter we can vary to control the signal-to-noise ratio of the task. These distractors are carefully selected from the same domain to be plausible but irrelevant.
    \item \textbf{Task Formulation:} The LLM is presented with the context axioms $A_{\text{context}}$, the theorem $T$, and the shuffled, numbered list of premises in $P_{\text{pool}}$. The prompt explicitly states that the theorem is \textbf{provable} from a subset of the provided pool and that its task is to identify the \textbf{precise minimal subset} of premises that are both necessary and sufficient for the proof.
\end{enumerate}

The LLM must return the list of indices corresponding to the premises in $P_{\text{minimal}}$. By jointly modulating the \texttt{proof depth} $d$ and the number of \texttt{distractors} $k$, we can create a fine-grained spectrum of tasks, from identifying short, obvious arguments in a clean context to pinpointing long, complex reasoning chains in a very noisy one.

\subsection{Task 3: Proof Graph Reconstruction}
\begin{figure}[h!]
    \centering 
    \tikzstyle{clause}  = [rectangle, rounded corners, draw=black, fill=blue!10, font=\footnotesize, align=center]
\tikzstyle{distractor}  = [rectangle, rounded corners, draw=black, fill=orange!10, font=\footnotesize, align=center]
\tikzstyle{axiom}   = [rectangle, rounded corners, draw=black, fill=green!20, font=\footnotesize, align=center]
\tikzstyle{target}  = [rectangle, rounded corners, draw=red!70, thick, fill=red!10, font=\footnotesize, align=center]
\tikzstyle{premise} = [rectangle, rounded corners, draw=orange!80, thick, fill=orange!10, font=\footnotesize, align=center]
\tikzstyle{panel_label} = [font=\bfseries]
\tikzstyle{box_label}   = [font=\tiny\itshape, text=black!70]
\tikzstyle{arrow}       = [->, >=Stealth, thick]
\tikzstyle{process_arrow} = [->, decorate, decoration={snake, amplitude=.3mm, segment length=1.5mm, post length=1mm}, thick, draw=orange!80]

\begin{minipage}{0.31\textwidth}
    \centering
    \vbox{
    \begin{tikzpicture}[xscale=0.8, yscale=0.8, node distance=1.0cm and 0.8cm]
        \node[panel_label] (label_a) at (0.5, 3.4) {A. Original Proof Graph};
        \node[axiom]  (ax1) at (-1.2, 2.0) {$C_1$};
        \node[axiom]  (ax2) at ( 0.0, 2.0) {$C_2$};
        \node[axiom]  (ax3) at ( 1.2, 2.0) {$C_3$};
        \node[axiom]  (ax4) at ( 2.4, 2.0) {$C_4$};
        \node[clause] (l1) at (-0.6, 0.7) {$C_5$};
        \node[clause] (l2) at ( 1.8, 0.7) {$C_6$};
        \node[target] (t)  at (0.6, -0.8) {$C_7$};
        \draw[arrow] (ax1) -- (l1); \draw[arrow] (ax2) -- (l1);
        \draw[arrow] (ax3) -- (l2); \draw[arrow] (ax4) -- (l2);
        \draw[arrow] (l1) -- (t);   \draw[arrow] (l2) -- (t);
        \node[fit=(label_a)(ax1)(ax2)(ax3)(ax4)(l1)(l2)(t), draw=black!20, rounded corners=3pt, inner sep=6pt] (panelAfit) {};
    \end{tikzpicture}
    }
\end{minipage}
\hfill
\begin{minipage}{0.25\textwidth}
    \centering
    \begin{tikzpicture}
        \node[panel_label] (label_b) at (0, 0.9) {B. Dissociate \& Shuffle};
        \draw[process_arrow] (-2,0) -- (2,0);
    \end{tikzpicture}
\end{minipage}
\hfill
\begin{minipage}{0.4\textwidth}
    \centering
    \begin{tikzpicture}
        \node[panel_label] (label_c) at (0, 3.4) {C. Task Formulation};
        \node[rectangle, draw, rounded corners, minimum width=4.8cm, minimum height=6.0cm] (llm_box) at (0,0) {};
        \node[align=left, text width=4.3cm, anchor=north] at ([yshift=-1.5mm]llm_box.north) {
            \textbf{Shuffled Clauses:}\\
            \footnotesize 1. Clause $C_3$\\ \footnotesize 2. Clause $C_7$ (Theorem)\\ \footnotesize 3. Clause $C_5$\\
            \footnotesize 4. Clause $C_1$\\ \footnotesize 5. Clause $C_6$\\ \footnotesize 6. Clause $C_4$\\
            \footnotesize 7. Clause $C_2$
        };
        \node[align=left, text width=4.3cm, anchor=south] at ([yshift=2mm]llm_box.south) {
            \textbf{Question:} Reconstruct the proof graph.\\
            \textbf{Answer Format:}\\
            \footnotesize 3 <- 4, 7\\
            \footnotesize 5 <- 1, 6\\
            \footnotesize 2 <- 3, 5};
        
        \draw[dashed, thin] ([yshift=-3mm]llm_box.west) -- ([yshift=-3mm]llm_box.east);
    \end{tikzpicture}
\end{minipage}    
    \caption{Generation pipeline for a Proof Graph Reconstruction task. A: select a complete, binary proof subgraph. B: dissociate structure and shuffle all clauses. C: the LLM receives the shuffled list and must output derivation steps that reconstruct the original DAG.}
        \label{Reconstruction Task}
\end{figure}
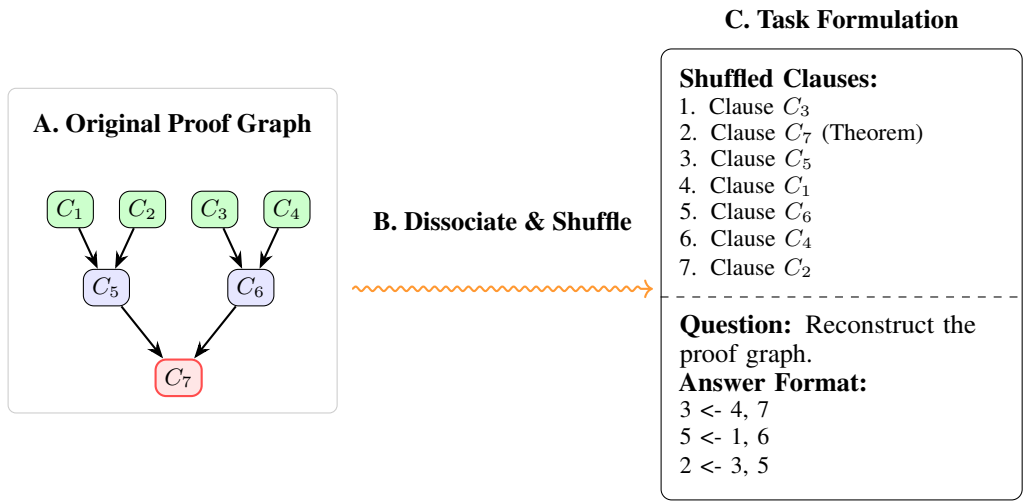
This task moves beyond local inference to test an LLM's capacity to understand and reconstruct the global, hierarchical structure of a complete proof.

\paragraph{Generation Process:}
\begin{enumerate}
    \item \textbf{Proof Subgraph Selection:} We select candidate proofs from our curated knowledge base that are well-suited for a structural reconstruction task. To ensure a clear and unambiguous dependency structure, we filter for proofs where each intermediate step is a \textbf{binary inference}, meaning every non-axiom clause is derived from exactly two parent clauses.
    \item \textbf{Complexity Control:} The difficulty of the task is controlled by the target \texttt{proof depth} $d$. We select proof subgraphs whose number of nodes (clauses) falls within a range corresponding to the typical size of a binary proof tree of depth $d$ and with at least one path to axioms of the exact length of the proof depth. This allows us to generate tasks with a predictable structural complexity.
    \item \textbf{Task Formulation:} Once a suitable proof subgraph for a theorem $T$ is selected, we provide the LLM with all its unique clauses as a shuffled, numbered list. The model must reconstruct the original derivation DAG by specifying the parent-child relationships for each non-axiom clause in the format \texttt{CHILD\_INDEX <- PARENT\_1\_INDEX, PARENT\_2\_INDEX}.
\end{enumerate}

\paragraph{Evaluation Metric:} The evaluation of the reconstructed graph is performed in two stages:
\begin{enumerate}
    \item \textbf{Structural Coherence:} First, we verify that the model's output forms a valid, acyclic directed graph where each derived node has exactly two distinct parent nodes from the provided list. Outputs that fail this basic structural check receive a score of zero.
    \item \textbf{Logical Soundness:} For structurally coherent graphs, we then assess their logical validity. We perform a semantic check on each proposed inference step (\texttt{CHILD <- PARENT\_1, PARENT\_2}) using Vampire to verify that the child clause is a correct logical consequence of its proposed parents. The final score is the proportion of logically sound inference steps in the reconstructed graph.
\end{enumerate}

This evaluation ensures that credit is awarded only for reconstructions that are both structurally well-formed and logically correct, providing a rigorous measure of an LLM's deep reasoning capabilities.

\section{Experiments and Discussion}

We conducted a series of experiments to validate the framework and probe the zero-shot reasoning limits of state-of-the-art Large Language Models. Our evaluation systematically measures the impact of logical complexity, task structure, and model scale on performance without any task-specific fine-tuning.

\subsection{Experimental Setup}

\paragraph{Dataset and Tasks.} We generated a benchmark of 3,000 problems, with 50 unique instances for each configuration. The benchmark covers five diverse TPTP domains : Algebra
(ALG), Fields (FLD), Geometry (GEO), Set Theory (SET), and Topology (TOP).
For Entailment Verification and Premise Selection, we created a fine-grained difficulty matrix. We defined four difficulty levels: \textbf{Level 1 (Easy)}, \textbf{Level 2 (Medium)}, \textbf{Level 3 (Hard)}, and \textbf{Level 4 (Very Hard)}. These levels are a function of both the proof depth ($d$) and the number of distractors/perturbations ($k$). For Entailment, $k$ was sampled from $\{2, 3, 4, 6\}$, while for Selection, $k$ was sampled from $\{2, 4, 6, 8\}$, with higher values of $d$ and $k$ corresponding to higher difficulty levels. For Proof Reconstruction, we limited our tests to proof depths of $d \in \{1, 2, 3, 4\}$, as deeper tasks proved intractable.

\paragraph{Models and Evaluation:} We evaluated three models representing different scales: \textbf{gpt-5-nano}, \textbf{gpt-5-mini}, and \textbf{gpt-5} with reasoning effort parameters set to low. All evaluations were conducted in a strict \textbf{zero-shot} setting, where models were given only the problem description and format instructions, without any in-context examples. Performance was measured using the metrics defined in Section 4 for each respective task.

\subsection{Results and Discussion}

\begin{figure}[h!]
    \centering 
    
    \includegraphics[width=1\textwidth]{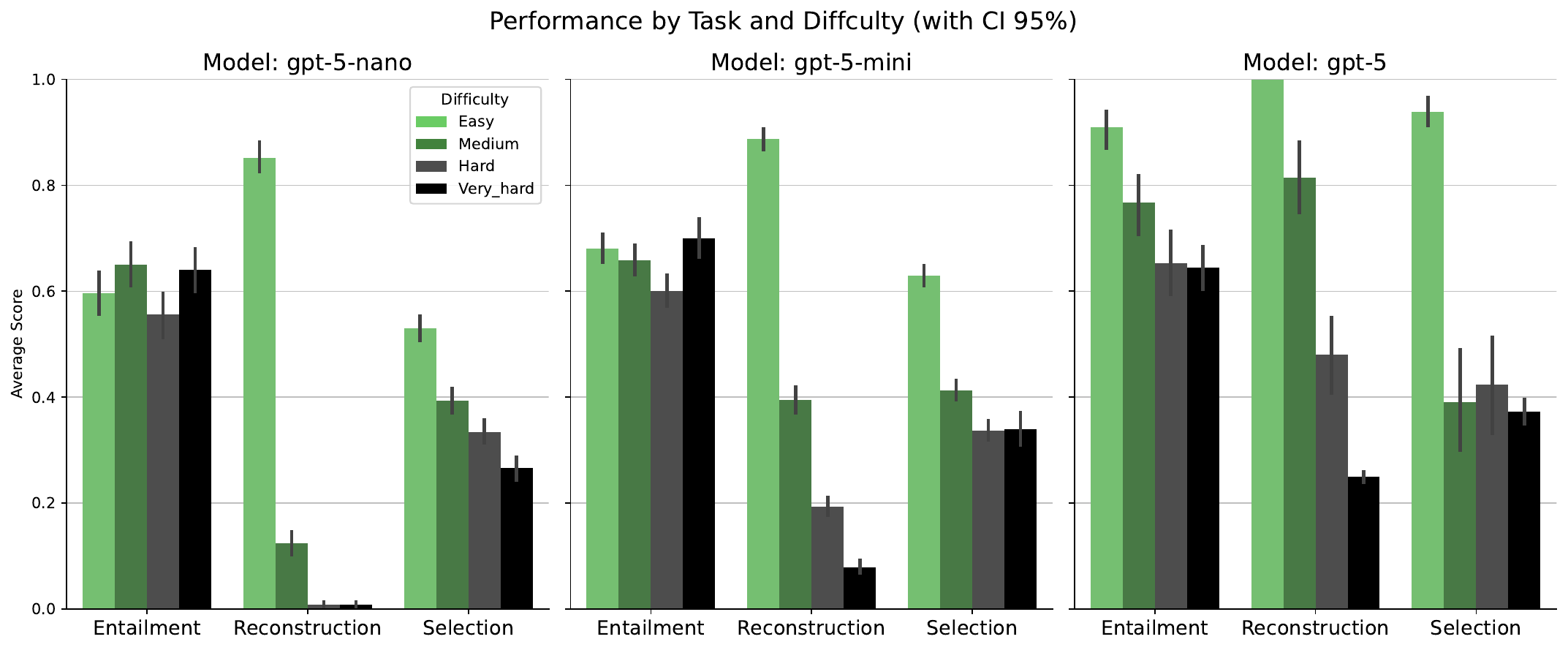}
    
    \caption{Performance comparison of three model scales on the three reasoning tasks, aggregated across all domains. The average score (y-axis) is plotted against four levels of task difficulty. Two key trends are apparent: (1) performance degrades for all models as difficulty increases, and (2) the structural reasoning required for the Reconstruction task proves significantly more challenging than other tasks, with smaller models failing almost completely.}

    \label{fig:difficulty}
\end{figure}

Our results, summarized in Figure~\ref{fig:difficulty}, reveal two primary findings regarding the zero-shot logical reasoning capabilities of current LLMs.

\textbf{First, performance systematically degrades with increased logical complexity.} Across all tasks and models, we observe a clear downward trend in scores as \texttt{proof depth} increases. This trend confirms that our framework's difficulty parametrization is effective and highlights that multi-step inference is a core bottleneck for LLMs. The degradation is particularly stark for the \textbf{Proof Reconstruction} task, where performance for smaller models collapses even at shallow depths. This suggests a fundamental weakness in global, structural reasoning; while models may handle local inference, they are largely incapable of assembling a coherent, multi-step deductive argument from scratch.

\begin{table*}[t!]
\centering
\caption{Model performance breakdown with difficulty level, reported as average score (\%).}
\label{tab:difficulty_breakdown}
\small
\begin{tabular}{@{}llcccc@{}}
\toprule
\textbf{Task Type} & \textbf{Model} & \textbf{Easy} & \textbf{Medium} & \textbf{Hard} & \textbf{Very Hard} \\
\midrule
\multirow{3}{*}{All} 
& gpt-5 & 93 & 68 & 54 & 44 \\
& gpt-5-mini & 72 & 48 & 37 & 41 \\
& gpt-5-nano & 65 & 38 & 29 & 35 \\

\bottomrule
\end{tabular}
\end{table*}

\textbf{Second, model scale is a key factor, but it is not a panacea for structural reasoning.} As expected, larger models consistently outperform smaller ones, especially on tasks with higher \texttt{proof\_depth} or more distractors. The \textbf{gpt-5} model, for instance, maintains a clear advantage on the most difficult problem sets. However, even this state-of-the-art model struggles immensely with the Reconstruction task. This indicates that while scaling improves performance on tasks requiring local deduction (Entailment) and noise filtering (Selection), it is not, by itself, sufficient to unlock robust, hierarchical planning abilities.

\paragraph{Exploring the Frontier: High-Effort Reasoning on Intractable Tasks.}
To further investigate the limits of structural reasoning, we conducted a small-scale exploratory study on high-difficulty Reconstruction tasks ($d=4$). We queried the largest state-of-the-art models using a high-effort "thinking" mode. The results (detailed in Appendix~\ref{sec:appendix}) show that despite generating extensive chains of thought, the model consistently failed to produce globally correct proof graphs. The reasoning often identified correct local dependencies but failed to integrate them into a valid, overarching structure. This reinforces our conclusion that the difficulty is not merely computational but stems from a core deficit in the architectural capabilities of current models for hierarchical, symbolic planning.

\textbf{In summary,} our experiments demonstrate that this framework provides a robust and granular benchmark for assessing zero-shot logical reasoning. The results clearly show that even state-of-the-art LLMs have profound weaknesses in multi-step, structural deduction, a limitation that is not fully overcome by simply increasing model scale. This highlights the critical need for training data and methods specifically designed to teach these complex reasoning skills, for which our framework can provide a potentially endless source of material.

\section{Conclusion and Future Work}

We have presented a novel framework for generating an endless supply of mathematically sound and structurally complex reasoning tasks. By leveraging the power of symbolic automated reasoning, our method ensures validity, provides fine-grained control over task difficulty, and covers a wide range of mathematical domains. Our preliminary results demonstrate that the framework can effectively probe the limitations of current LLMs, showing a marked decline in performance as logical depth increases, particularly in tasks requiring global structural reasoning.

This work opens several exciting avenues for future research.

\paragraph{Training for Generalization: From Symbolic to Natural Language.}
Our primary research direction is to use the datasets generated by our framework for fine-tuning Large Language Models. This will allow us to test a central hypothesis: are symbolic reasoning skills transferable to natural language? We plan to evaluate models fine-tuned on our data on existing benchmarks of mathematical word problems to determine if mastering the underlying logical structure improves general reasoning capabilities.

\paragraph{Scaling Complexity with Iterative Saturation.}
To generate even more complex and diverse problems, we will implement an iterative saturation loop, an approach inspired by AGInT \citep{puzis2006}. In this setup, the most "interesting" theorems from one generation round would be fed back as new axioms for the next. This would allow us to explore the deductive closure of a theory more deeply. However, we note that our current single-pass saturation method is already more than sufficient to generate tasks that challenge even the most powerful state-of-the-art models.

\paragraph{Expanding the Logical Horizon to FOL.}
Finally, we plan to extend the framework to support full first-order logic (FOL). Since our backend provers are already optimized for FOL, this is a natural next step that will greatly increase the expressiveness of our tasks. We are releasing our open-source code and generated datasets to facilitate this research, with the ultimate goal of fostering a new generation of AI systems that can reason with mathematical rigor.

\section{Limitations} 
Limitations and Design Choices. Our framework makes deliberate design choices that trade generality for rigor. First, we work exclusively in CNF to ensure complete control over logical validity—while this abstracts away natural language complexity, it allows us to isolate pure reasoning capabilities without confounding factors. This is a feature, not a bug: by establishing baseline performance on clean logical tasks, we can better understand where reasoning fails before adding linguistic complexity.

Second, our evaluation focuses on the GPT family as a representative sample of state-of-the-art models. While broader evaluation would be valuable, the consistent architecture allows us to isolate the effect of scale on reasoning depth. The framework is model-agnostic and can be applied to any LLM.

Third, the current scale (3,000 problems) represents a proof of concept. The generative nature of our framework means it can produce arbitrary amounts of data on demand—we deliberately kept the initial evaluation focused to establish clear trends rather than pursue scale for its own sake.

\clearpage
\small
\bibliography{custom}
\bibliographystyle{plainnat}

\section{Appendix}
\label{sec:appendix}

\subsection{Prompt Examples for Generated Tasks}

\subsubsection{Task 1: Conjecture Entailment Verification}

The theorem used in this example demonstrates a fundamental property of set theory.
\begin{description}
    \item[Formal Clause:] \texttt{(disjoint(X1,complement(X2))|\textasciitilde{}subset(X1,X2))}
    \item[Summary:] \textit{A set fully contained within another has no elements outside of it.}
\end{description}

Below is a full example of the prompt provided to the LLM for the entailment task.

\begin{lstlisting}[style=PromptStyle, caption={Prompt for the Entailment Task.}, label={lst:entailment}]
You will be given a logical entailment problem in three parts.

PART 1: CONTEXT
The following are general axioms from the domain of **Algebra**. 
They provide definitions and background theory. 
**Do NOT use them directly in the proof.**

- cnf(disjoint2,axiom,(disjoint(X1,X2)|member(f23(X1,X2),X1)))
- cnf(complement1,axiom,(~member(X1,complement(X2))|~member(X1,X2)))
- cnf(disjoint3,axiom,(disjoint(X1,X2)|member(f23(X1,X2),X2)))
- cnf(subset1,axiom,(member(X3,X2)|~subset(X1,X2)|~member(X3,X1)))


PART 2: THE SPECIFIC PROBLEM
Your task is to evaluate the following specific entailment claim.

**Premises to use:**
- (disjoint(X1,complement(X2))|~member(f23(X1,complement(X2)),X2))
- (subset(image(X1,domain_of(X2)),X3)|~disjoint(X2,universal_set))
- (associative(X1,X2)|~disjoint(X1,X3)|~member(f35(X1,X2),X3))
- (disjoint(X1,X2)|member(f23(X1,X2),X3)|~subset(X1,X3))


**Conclusion to prove:**
(disjoint(X1,complement(X2))|~subset(X1,X2))


PART 3: YOUR TASK
Based **only** on the 'Premises to use', does the 'Conclusion to prove' logically follow?
Answer with a single word: `True` or `False`.

--- EXPECTED OUTPUT --- 'True'
\end{lstlisting}

\subsubsection{Task 2: Minimal Premise Selection}

This task uses a non-trivial theorem related to the Axiom of Infinity, linking infinite-like sets to the structure of the universe.
\begin{description}
    \item[Formal Clause:] \texttt{(\textasciitilde{}inductive(X1)|\textasciitilde{}subclass(X1,complement(X2))|\textasciitilde{}subclass(universal\_class,X2))}
    \item[Summary:] \textit{If an infinite-like set exists outside of another set, then that latter set cannot be the entire universe.}
\end{description}

Below is a full example of the prompt provided to the LLM for the selection task.

\begin{lstlisting}[style=PromptStyle, caption={Prompt for the Premise Selection Task.}, label={lst:selection}]
You are a mathematical logic assistant. 
Your task is to identify a minimal set of premises sufficient for a proof.

## General Context
The problem is set in the domain of: **Set Theory**.
The following are the fundamental axioms of this domain. 
They provide general context. **Do not use them in the proof itself.**
Fundamental Axioms:
- cnf(equal_implies_subclass2,axiom,(subclass(X2,X1)|X1!=X2))
- cnf(complement1,axiom,(~member(X1,complement(X2))|~member(X1,X2)))
- cnf(inductive1,axiom,(member(null_class,X1)|~inductive(X1)))
- cnf(omega_is_inductive2,axiom,(subclass(omega,X1)|~inductive(X1)))
- cnf(omega_is_inductive1,axiom,(inductive(omega)))
- cnf(not_subclass_members1,axiom,(member(not_subclass_element(X1,X2),X1)|subclass(X1,X2)))
- cnf(regularity1,axiom,(X1=null_class|member(regular(X1),X1)))
- cnf(subclass_implies_equal,axiom,(X1=X2|~subclass(X1,X2)|~subclass(X2,X1)))
- cnf(subclass_members,axiom,(member(X3,X2)|~subclass(X1,X2)|~member(X3,X1)))
- cnf(not_subclass_members2,axiom,(subclass(X1,X2)|~member(not_subclass_element(X1,X2),X2)))
- cnf(class_elements_are_sets,axiom,(subclass(X1,universal_class)))

## Task
Your goal is to prove the following theorem:
**Theorem:**
`(~inductive(X1)|~subclass(X1,complement(X2))|~subclass(universal_class,X2))`

Below is a numbered pool of potential premises.
Your task is to identify the **minimal subset** of numbers from this pool 
whose corresponding statements are **sufficient on their own** to prove the theorem.

**Pool of Premises:**
1. (subclass(complement(X1),X2)|~subclass(universal_class,X1))
2. (null_class=X1|null_class=X2|~inductive(unordered_pair(X2,X1)))
3. (~inductive(X1)|~subclass(X2,null_class)|~subclass(X1,X2))
4. (member(omega,X1)|intersection(X2,X1)!=universal_class)

### Question
Which is the smallest set of numbered premises from the pool that is sufficient to prove the theorem,
without using the fundamental axioms from the context?

### Response Format
Your answer must be **only** a list of numbers, sorted in increasing order. For example: `[2, 5, 8]`.

--- EXPECTED OUTPUT ---[1, 3, 7]
\end{lstlisting}

\subsubsection{Task 3: Proof Graph Reconstruction}

This example features a complex theorem from topology, illustrating a consistency constraint within the formal system.
\begin{description}
    \item[Formal Clause:] \parbox[t]{0.8\textwidth}{
    \texttt{(subset\_sets(X1,X2) | \\
    \textasciitilde{}element\_of\_collection(union\_of\_members(top\_of\_basis
    \text{}(top\_of\_basis(subspace\_topology(X2,X3,X1))))),X4) | \\
    \textasciitilde{}element\_of\_set(X5,intersection\_of\_members(X4)))
    }}
    \item[Summary:] \textit{To logically define a 'subspace' structure, the object must first be a 'subset'.}
\end{description}

Below is a full example of the prompt provided to the LLM for the reconstruction task.

\begin{lstlisting}[style=PromptStyle, caption={Prompt for the Proof Graph Reconstruction Task.}, label={lst:reconstruction}]
Your task is to reconstruct the dependency graph of a mathematical proof from the domain of **Topology**.

The proof graph concludes with the theorem: 
`(subset_sets(X1,X2)|~element_of_collection(union_of_members(top_of_basis(top_of_basis(subspace_topology(X2,X3,X1)))),X4)|~element_of_set(X5,intersection_of_members(X4)))`

## Proof Context & Rules
This proof was generated by using the **Superposition Calculus** (which includes rules like Resolution and Paramodulation).

Therefore, the proof has the following properties:
- **Starting Points:** Some clauses in the list are starting points (axioms) and are not derived from other clauses.
- **Derived Clauses:** Every other clause is derived from exactly **two** parent clauses from the list.
- **Clause Reuse:** A single clause can be used as a parent in multiple derivation steps.

## Your Task
Given the rules above, reconstruct the proof from the following shuffled list of clauses.
Identify the derivation for every clause that is not a starting point.

**Shuffled Clauses:**
1. (element_of_set(X3,f10(X2,X1,X3))|~element_of_collection(X1,top_of_basis(X2))|~element_of_set(X3,X1))
2. (subset_sets(X1,X2)|~element_of_set(X3,union_of_members(top_of_basis(subspace_topology(X2,X4,X1)))))
3. (element_of_set(X1,f1(X2,X1))|~element_of_set(X1,union_of_members(X2)))
4. (element_of_collection(f1(X2,X1),X2)|~element_of_set(X1,union_of_members(X2)))
5. (element_of_set(X1,union_of_members(X2))|~element_of_set(X1,X3)|~element_of_collection(X3,X2))
6. (subset_sets(X4,X2)|~element_of_collection(X1,subspace_topology(X2,X3,X4)))
7. (element_of_set(X1,union_of_members(X2))|~element_of_set(X1,union_of_members(top_of_basis(X2))))
8. (element_of_collection(f10(X2,X1,X3),X2)|~element_of_collection(X1,top_of_basis(X2))|~element_of_set(X3,X1))
9. (subset_sets(X1,X2)|~element_of_collection(f1(X3,X4),top_of_basis(subspace_topology(X2,X5,X1)))|~element_of_set(X4,union_of_members(X3)))
10. (element_of_set(X1,X3)|~element_of_set(X1,intersection_of_members(X2))|~element_of_collection(X3,X2))
11. (subset_sets(X1,X2)|~element_of_collection(union_of_members(top_of_basis(top_of_basis(subspace_topology(X2,X3,X1)))),X4)|~element_of_set(X5,intersection_of_members(X4)))
12. (element_of_set(X1,union_of_members(X2))|~element_of_collection(f10(X3,X4,X1),X2)|~element_of_collection(X4,top_of_basis(X3))|~element_of_set(X1,X4))
13. (element_of_set(X1,union_of_members(X2))|~element_of_collection(f1(X3,X1),top_of_basis(X2))|~element_of_set(X1,union_of_members(X3)))
14. (element_of_set(X1,union_of_members(X2))|~element_of_collection(X3,top_of_basis(X2))|~element_of_set(X1,X3))
15. (subset_sets(X1,X2)|~element_of_collection(X3,top_of_basis(subspace_topology(X2,X4,X1)))|~element_of_set(X5,X3))
16. (element_of_set(X1,union_of_members(X2))|~element_of_collection(union_of_members(top_of_basis(X2)),X3)|~element_of_set(X1,intersection_of_members(X3)))

## Required Output Format
- List **only** the derivation steps.
- Each step must be on a new line.
- Use the exact format `CHILD <- PARENT_1, PARENT_2`. Example: `5 <- 2, 4`.
- All clauses from the list must be used in the final structure.
- No explanations, comments, or extra text.

--- EXPECTED OUTPUT ---
11 <- 2, 16
12 <- 1, 5
13 <- 3, 14
14 <- 8, 12
15 <- 6, 8
16 <- 7, 10
2 <- 4, 9
7 <- 4, 13
9 <- 3, 15
\end{lstlisting}

\subsection{Visual Comparison of Proof Graph Reconstructions}
\label{sec:reconstruction_comparison}

This section provides a detailed comparison of the proof graph reconstruction capabilities of various state-of-the-art models. The task is to reconstruct the dependency graph for the topology theorem on the logical consistency of subspace definitions, as detailed in Section~\ref{sec:appendix}. Each model was prompted with the full context and allowed a reasoning time of approximately 15 minutes to generate its solution.

For each model, we present a single, structured visualization that overlays the model's reconstruction onto the ground truth graph. The edges are color-coded for immediate analysis:
\begin{itemize}
    \item \textcolor{green!50!black}{\textbf{Green edges}} are correct dependencies, present in both the ground truth and the model's output (True Positives).
    \item \textcolor{gray}{\textbf{Dashed gray edges}} are correct dependencies that the model failed to identify (False Negatives).
    \item \textcolor{red!80!black}{\textbf{Red edges}} are incorrect dependencies that were hallucinated by the model (False Positives).
\end{itemize}

\subsubsection{Model: \texttt{Gemini 2.5 pro(max thinking)}}

\begin{figure}[h!]
    \centering
    \includegraphics[width=0.8\textwidth]{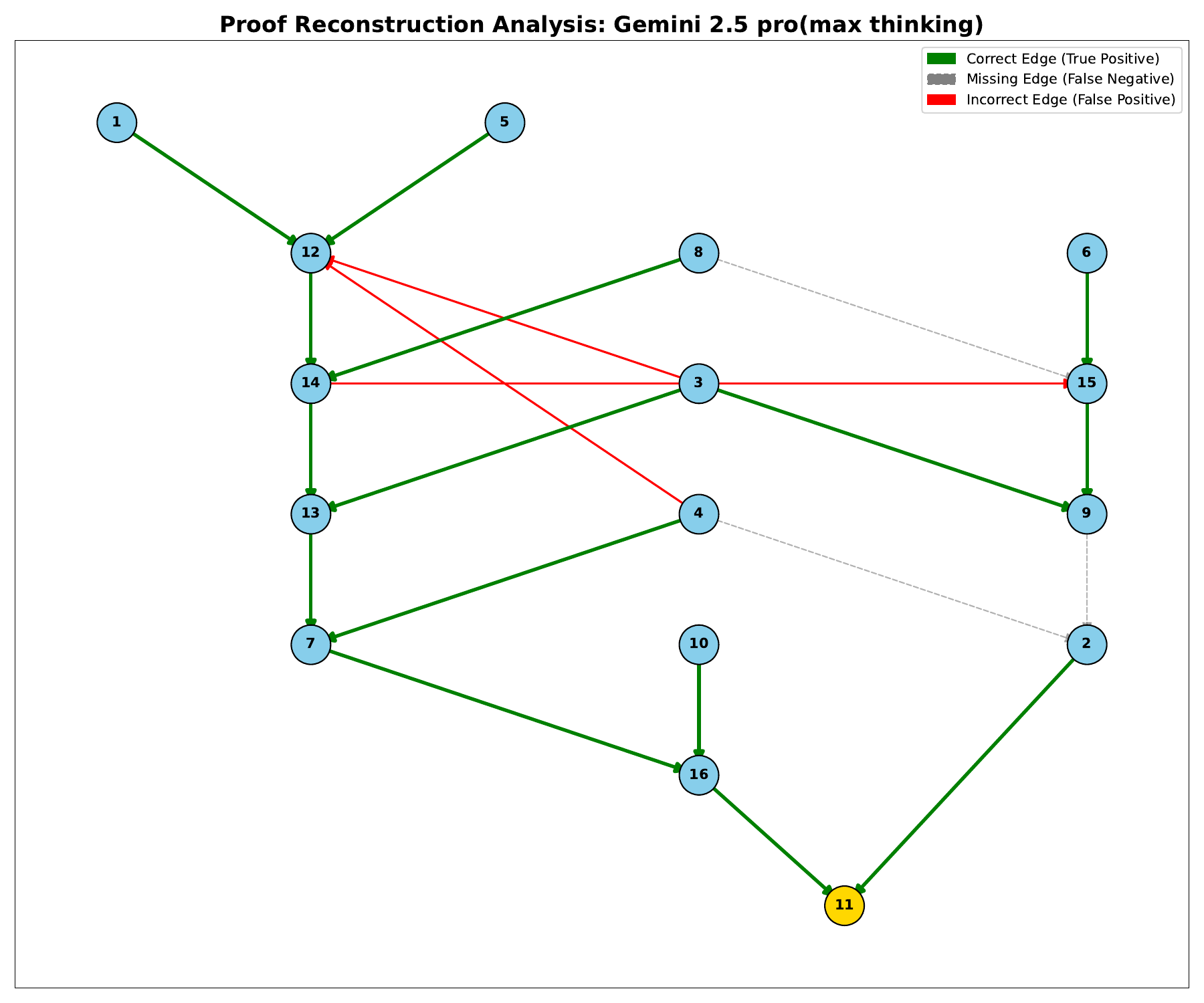}
    \caption{Proof graph reconstruction analysis for \texttt{Gemini 2.5 pro}.}
    \label{fig:gemini_comparison}
\end{figure}

\paragraph{Qualitative Analysis:} This model demonstrates strong performance, reconstructing most of the dependencies correctly. It makes a few specific errors, including hallucinating several dependencies for clause 12 and using an incorrect parent for clause 15. It also fails to reconstruct the derivation for clause 2.

\paragraph{Model Output:}
\begin{lstlisting}[style=ModelOutputStyle]
12 <- 4, 9
7 <- 4, 13
9 <- 3, 15
11 <- 2, 16
12 <- 1, 5
13 <- 3, 14
14 <- 8, 12
15 <- 6, 14
16 <- 7, 10
\end{lstlisting}

\subsubsection{Model: \texttt{Claude-4.1-Opus(thinking)}}

\begin{figure}[h!]
    \centering
    \includegraphics[width=0.8\textwidth]{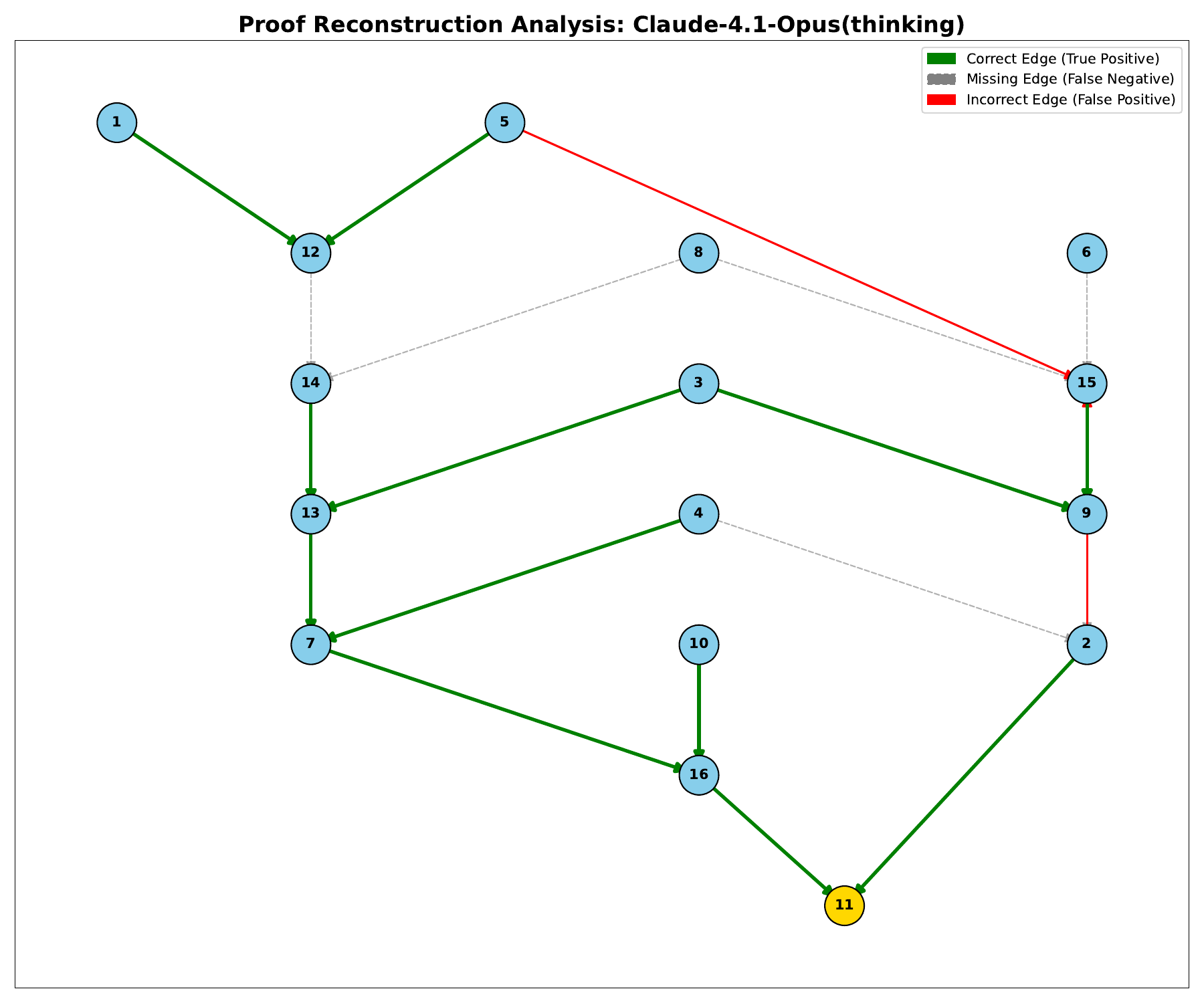}
    \caption{Proof graph reconstruction analysis for \texttt{Claude-4.1-Opus}.}
    \label{fig:claude_comparison}
\end{figure}

\paragraph{Qualitative Analysis:} The model correctly identifies a significant portion of the proof structure, particularly the main path leading to the conclusion. However, it hallucinates two incorrect dependencies for clause 15 and fails to derive several key intermediate clauses, notably clauses 14 and 2.

\paragraph{Model Output:}
\begin{lstlisting}[style=ModelOutputStyle]
7 <- 13, 4
9 <- 15, 3
11 <- 2, 16
12 <- 5, 1
13 <- 14, 3
15 <- 2, 5
16 <- 10, 7
\end{lstlisting}
\clearpage

\subsubsection{Model: \texttt{Deepseek V3.1}}

\begin{figure}[h!]
    \centering
    \includegraphics[width=0.8\textwidth]{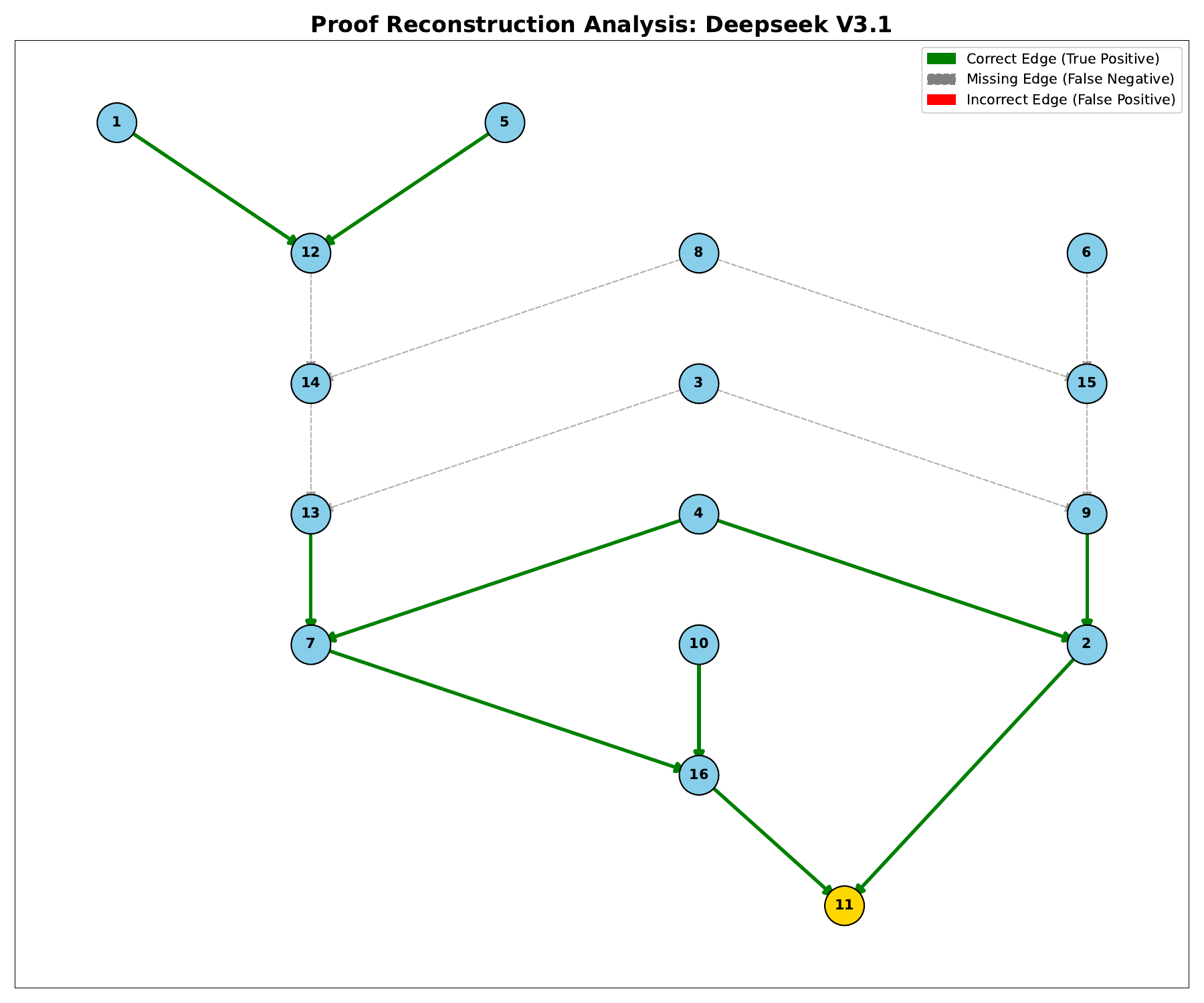}
    \caption{Proof graph reconstruction analysis for \texttt{Deepseek V3.1}.}
    \label{fig:deepseek_comparison}
\end{figure}

\paragraph{Qualitative Analysis:} This model produces a highly precise but incomplete graph. Every dependency it identifies is correct (high precision), but it fails to reconstruct a large portion of the proof, leaving several clauses (9, 13, 14, 15) without inferred parents, indicating a significant failure in recall.

\paragraph{Model Output:}
\begin{lstlisting}[style=ModelOutputStyle]
2 <- 9,4
7 <- 13,4
11 <- 16,2
12 <- 5,1
16 <- 10,7
\end{lstlisting}
\clearpage
\subsubsection{Model: \texttt{Gpt-5(high)}}

\begin{figure}[h!]
    \centering
    \includegraphics[width=0.8\textwidth]{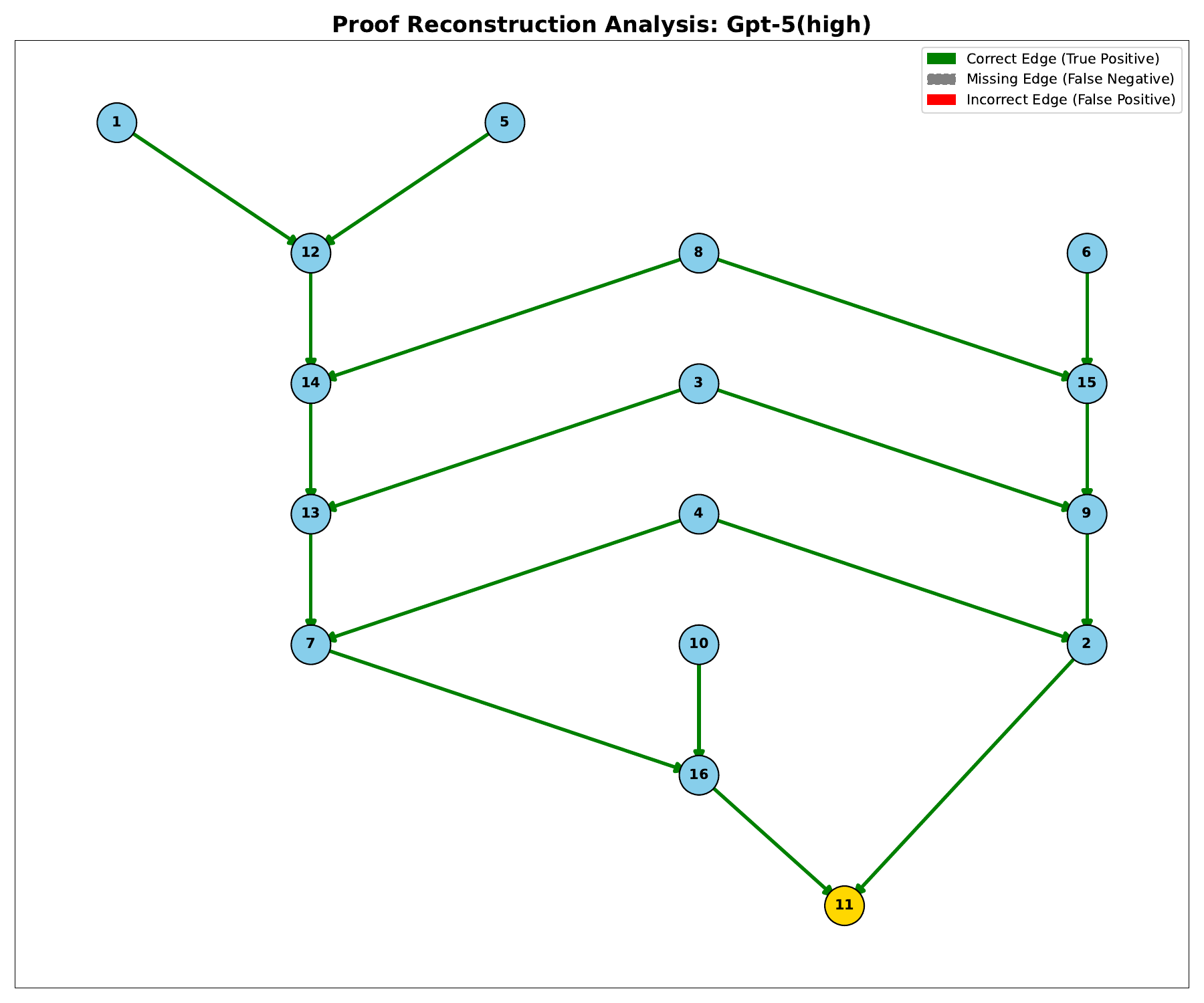}
    \caption{Proof graph reconstruction analysis for \texttt{Gpt-5(high)}.}
    \label{fig:gpt5_comparison}
\end{figure}

\paragraph{Qualitative Analysis:} The model provides a perfect reconstruction of the proof graph, identifying all 18 dependencies correctly without any missing or hallucinated edges. The generated structure is identical to the ground truth, demonstrating a complete and accurate understanding of the proof's logical flow.

\paragraph{Model Output:}
\begin{lstlisting}[style=ModelOutputStyle]
12 <- 5, 1
15 <- 6, 8
14 <- 12, 8
9 <- 15, 3
13 <- 14, 3
2 <- 9, 4
7 <- 13, 4
16 <- 10, 7
11 <- 2, 16
\end{lstlisting}
\clearpage
\end{document}